\documentclass{article}

\usepackage{microtype}
\usepackage{graphicx}
\usepackage{subfigure}
\usepackage{booktabs} % for professional tables
\usepackage{bold-extra}
\usepackage{xcolor}
\usepackage{amsmath}
\usepackage{amsfonts}
\usepackage{comment}

\definecolor{megha-purple}{rgb}{0.55, 0.24, 1}

\newcommand{\model}[0]{\texttt{\textbf{\textcolor{orange}{CORGI}}}}

\usepackage{comment}
\usepackage[accepted]{icml2023}
\usepackage[hidelinks]{hyperref}
\hypersetup{
    colorlinks=true,
    linkcolor=olive,
    filecolor=magenta,      
    urlcolor=olive,
    citecolor=olive,
}

\icmltitlerunning{Generating Language Corrections for Teaching Physical Control Tasks}

\begin{document}

\twocolumn[
\icmltitle{Generating Language Corrections for Teaching Physical Control Tasks}

\icmlsetsymbol{equal}{*}

\begin{icmlauthorlist}
\icmlauthor{Megha Srivastava}{to}
\icmlauthor{Noah Goodman}{to,goo}
\icmlauthor{Dorsa Sadigh}{to}
\end{icmlauthorlist}

\icmlaffiliation{to}{Department of Computer Science, Stanford University}
\icmlaffiliation{goo}{Department of Psychology, Stanford University}

\icmlcorrespondingauthor{Megha Srivastava}{megha@cs.stanford.edu}

\icmlkeywords{Machine Learning, ICML}

\vskip 0.3in
]

\printAffiliationsAndNotice{} % otherwise use the standard text.
\begin{abstract}
AI assistance continues to help advance applications in education, from language learning to intelligent tutoring systems, yet current methods for providing students feedback are still quite limited. Most automatic feedback systems either provide binary correctness feedback, which may not help a student understand \textit{how} to improve, or require hand-coding feedback templates, which may not generalize to new domains. This can be particularly challenging for physical control tasks, where the rich diversity in student behavior and specialized domains make it challenging to leverage general-purpose assistive tools for providing feedback. 
We design and build \model{}, a model trained to generate language corrections for physical control tasks, such as learning to ride a bike. \model{} takes in as input a pair of student and expert trajectories, and then generates natural language corrections to help the student improve. We collect and train \model{} over data from three diverse physical control tasks (drawing, steering, and joint movement). Through both automatic and human evaluations, we show that \model{} can (i) generate valid feedback for novel student trajectories, (ii) outperform baselines on domains with novel control dynamics, and (iii) improve student learning in an interactive drawing task.

% three complex physical control tasks
\end{abstract}
\section{Introduction}
\label{introduction}
In our daily lives, we need to learn a variety of physical control tasks (e.g. driving a car or athletic sports) that benefit from receiving feedback of different modalities, such as visual demonstrations or haptic guidance. One of the most general forms of corrective feedback, however, is natural language -- a person learning how to ride a bike can easily understand what \textit{``make a sharper left turn''} means, even if they are unfamiliar with the specific control dynamics of the task. While recent works have focused on learning control policies that incorporate natural language feedback from users \citep{broad2017realtime, cui2023lilac, sharma2022correcting}, few have considered the reverse direction of automatically generating language corrections to provide to human users. Such corrections can be useful for enhancing human-AI interaction in decision making contexts \citep{lai2019explanations}, improving interactive data collection \citep{gandhi2022eliciting, gopalan2022negativelfd}, and more generally teaching humans how to perform physical control tasks such as for rehabilitation, flying an aircraft, or operating surgical robots.
% control simulators or rehabilitation technologies 
\cite{hays2009flight, maciejasz2014rehab, srivastava2022teaching, yu2022coach, schrum2022reciprocal}.

How do humans typically provide natural language feedback? Consider a parent who is teaching their child how to ride a bike. One form of corrective feedback they may provide are general, vague utterances (e.g. \textit{``that was okay, try again"}) that  provide positive or negative reinforcement, but may not be very informative on \textit{how} to improve. On the other extreme, the parent may provide precise feedback (e.g. \textit{``wider grip on the handle-bars"}) that clearly conveys how the child should adjust their behavior, but requires access to domain-specific information such as referring to handle bars, which is only applicable to the setting of teaching how to ride a bike. This results in a trade-off between helpfulness, or the ability to provide sufficient information to help a student improve, of corrections and their generality, or ability to be understood and conveyed across different settings.

In fact, existing works on automatic feedback generation in domains such as programming and language learning reflect this trade-off \citep{settles2020duolingo, liu2022feedback}. Some systems provide simple binary feedback (e.g. whether a program ran successfully), which may not be very helpful to the student, while others require hand-coded, templates (e.g. grammar checking) that lack generality.  
Due to the rich diversity of physical control tasks and variation in ways a student might under-perform, we seek to strike a balance by learning to generate helpful comparative corrections (e.g. \textit{``brake sooner"}) that can also generalize to novel trajectories within the same control space. To achieve this, we choose to leverage the expressive capabilities of language models (LMs), driven by the key insight that LMs may encode physical conceptual spaces that are isomorphic across the variety of environments, states, and action spaces that exist across different physical control tasks \cite{patel2022conceptual}.
%which can be a burden, hand-coded, requires domain vocab knowledge

Concretely, we design and build \model\footnote{\model{}: The acronym stands for natural language \textbf{cor}rections \textbf{g}eneration for \textbf{i}nstruction.}, a model trained to generate corrections in natural language based on three physical control tasks of drawing, driving a car in simulation, and dancing. These three tasks exhibit different control spaces such as the 2D x-y position on a surface, steering and acceleration, and skeleton joint motion, which in turn require \model{} to develop a general understanding of physical concepts. At test time, \model{} takes in as input a pair of student and expert trajectories, and generates a correction in natural language to help the student better match the expert's performance. Specifically, \model{} consists of a trainable trajectory encoder that learns to map student and expert trajectories to prompts that can be used as inputs to a frozen LM to generate feedback with,  thus keeping the more general representations of language encoded by the LM fixed.  Through both automatic and human evaluations, we show that \model{} can (i) generate valid feedback for novel student trajectories, (ii) outperform baselines on domains with novel control dynamics, and (iii) improve student learning in an interactive drawing task. Thus, in addition to introducing the task of generating natural language feedback to humans for physical control tasks, our contributions include:
\begin{enumerate}
\item A dataset of 2k crowdsourced corrections collected across (student, expert) trajectories from a diverse set of control tasks (drawing, steering, and joint motion).

    \item \model, our model trained to generate corrective feedback in natural language for these three tasks.

    \item A comprehensive evaluation of the ability of \model{} to generalize to novel student trajectories and domains that share the same control space.

    \item Two human subject user studies assessing both preference and the helpfulness of generated feedback in helping users improve drawing.

   % \item A detailed exploratory investigation into the zero-shot grounding capabilities of large-scale LMs for \textit{physical} attributes (weight, speed, angle, height). %which may be of additional interested to the NLP community?
\end{enumerate}

We will release all data, model checkpoints, code, and user study infrastructure to aid future work at \url{https://github.com/Stanford-ILIAD/corgi}. 

%Concretely, we design and build \model, a model trained for natural language \textbf{cor}rections \textbf{g}eneration for the \textbf{i}nstruction of three physical control tasks: \textsc{drawing}, \textsc{steering} a vehicle, and body \textsc{movement}. 

\section{Related Works}
While recent works have explored generating \textit{comparative}  descriptions, such as language descriptions of  distribution shifts \citep{zhong2022diff} and relative image captions \citep{mirchandani2022fadvlp}, we are the first to explore this for physical control tasks, as well as with an educational focus.
%https://arxiv.org/pdf/2201.12323.pdf

\label{related}
\noindent{\textbf{Language in Multimodal Tasks}}
Several  works have leveraged advances in LMs and multimodal models to improve human interaction across physical control tasks.  For example, Google's SayCan leverages LMs to break down language instructions into executable skills, providing users flexibility in receiving robotic assistance for complex, long-horizon tasks \citep{ahn2022saycan}. Others have explored using language to adjust robot plans with constraints or specify subgoals \citep{sharma2022correcting, karamcheti2021lila, cui2023lilac}. Finally, \cite{tevet2022motionclip} recently introduced MotionCLIP, a transformer-based auto-encoder that shows exciting text-to-motion capabilities like adjusting motion sequences for novel styles (e.g. \textit{``run away hysterically''}). 

Another multimodal task closely related to ours is image (or video) captioning, where large-scale multimodal models have achieved state-of-the-art performance on classic benchmarks such as MSCOCO \cite{alayrac2022flamingo, lin2014microsoft}. Furthermore, \citet{tsimpoukelli2021frozen} achieve strong performance on captioning tasks by only training a visual encoder to output a prompt for a frozen LM, motivating our approach for \model{}. 

\noindent{\textbf{Language in Education}}
A few works have studied the role of language descriptions and feedback in educational settings. 
\citet{chopra2019ratchet} show that language can reduce time in communicating concepts to a student, \citet{sumers2020show} find in a cooperative teaching game that language helps communicate more nuanced concepts than other feedback forms like demonstrations, and \citet{ruian2019quiz} demonstrate that interactive dialogue-based agents can improve student learning. However, these works largely focus on understanding the role of language in pedagogical settings, not automatically generating language feedback. 

\noindent{\textbf{Language in Physical Interaction Datasets}}
 Large-scale datasets of language paired with physical interactions have enabled further understanding of physical reasoning, as well as inspired progress on novel interactive control tasks. For example, 
\citet{ji2022tangrams} built a rich-annotated dataset of tangram puzzles to study the abstract visual reasoning capabilities of multi-modal models,  \citet{wong2022language} show  how to leverage annotations in the CLEVR dataset \citep{johnson2017clevr} to improve generalization on spatial relationship tasks and \citet{lynch2021play} show that ``play'' data annotations enable strong zero-shot language conditioning for robotic tasks.  
To the best of our knowledge, we are the first to collect corrections over pairwise trajectories, providing insight into how people reason about physical comparisons.

% http://www.columbia.edu/~mvp19/ETF/Feedback.pdf
\section{Generating Corrective Feedback}
\begin{figure*}
\vskip 0.2in
\begin{center}
\centerline{\includegraphics[width=\linewidth]{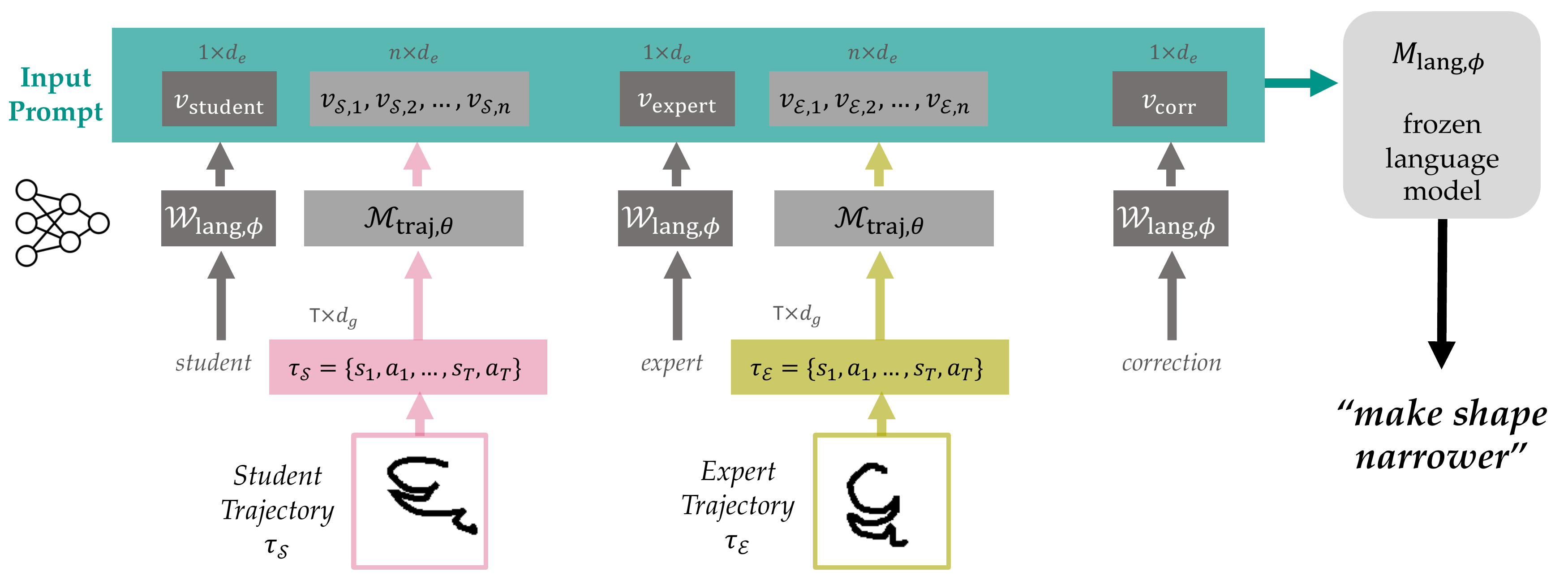}}
\caption{Overview of \model{} at test time. Trajectories $\tau_\mathcal{S}$, $\tau_\mathcal{E}$, from a student and an expert respectively, are mapped by a learned trajectory encoder $M_{\text{traj},\theta}$ to vectors of the same dimension as the output of the frozen language model $M_{\text{lang},\phi}$'s embedding layer ($W_{\text{lang},\phi}$). The resulting output vectors are stitched together with the embeddings corresponding to vocabulary words ``student'', ``expert'', and ``correction'' in order to create the input prompt sent to the  $M_{\text{traj},\theta}$, from which we then generate a correction.}
\label{fig:model-overview}
\end{center}
\vskip -0.2in
\end{figure*}
We now formalize generating corrective feedback in an educational setting, where the goal is to generate corrections from the set of possible natural language utterances $u \in \mathcal{U}$ that are \textit{comparative} with respect to some expert behavior. Consider a target physical control task $g$ (e.g. riding a bike), a student  $\mathcal{S}$ (e.g a child learning to ride a bike), and an expert $\mathcal{E}$ (e.g. their parent who can already perform this task). We can treat $g$ as a standard Markov decision process (MDP) $<S, A, f, R, T>$ with finite horizon $T$, reward function $R: S \times A \rightarrow \mathbb{R}$ 
over state ${S}$ and action ${A}$ spaces, and a deterministic transition function $f: S \times A \rightarrow S$ that maps a particular state and action pair $s_t, a_t$ at time step $t$ to a new state $s_{t+1}$. We can then define a trajectory $\tau$ as a sequence of state and action pairs $\{s_1,a_1, \dots, s_T,a_T\}$, and can collect trajectories from both the student ($\tau_\mathcal{S}$) and the expert ($\tau_\mathcal{E}$). Under this setting, we now formalize the goal of generating corrective feedback $u$ for the student $\mathcal{S}$.

\subsection{Problem Statement}
Effective feedback should reduce discrepancies between a student learner's current understanding and performance of a task and that of an expert teacher \citep{hattie2007feedback}. Therefore, good corrections should not only accurately identify such discrepancies, but also be sufficiently \textit{helpful} for the student to improve. We thus assess a correction $u$ by measuring the degree it reduces the gap between the student $\mathcal{S}$'s and expert $\mathcal{E}$'s performance on task $g$.

Concretely, let $\pi_{\mathcal{S},g}^k$  represent the student policy for task $g$ at time $k$ and  $\pi_{\mathcal{E},g}$ represent a fixed expert policy for task $g$. From these policies, we can collect trajectory rollouts  $\tau_\mathcal{S}^{{g,k}}$ and 
$\tau_\mathcal{E}^{g}$, respectively. Furthermore, let $\mathcal{L}$ be a task-dependent loss function that measures the discrepancy between two trajectories. A corrective feedback utterance $u_k$ provided at timestep $k$ may result in the student updating their policy from $\pi_{\mathcal{S},g}^k$ to $\pi_{\mathcal{S},g}^{k+1}$, and so the optimal corrective feedback would be a $u_k$ that minimizes the expression:
\setlength{\belowdisplayskip}{0pt} \setlength{\belowdisplayshortskip}{0pt}
\setlength{\abovedisplayskip}{3pt} \setlength{\abovedisplayshortskip}{3pt}
\begin{equation}
\label{eq:loss}
    \min_{u_k}\:\mathcal{L}
        (\tau_\mathcal{S}^{{g,k+1}}(u_k), \tau_\mathcal{E}^{g}) - 
        \mathcal{L}(\tau_\mathcal{S}^{{g,k}}, \tau_\mathcal{E}^{g})
\end{equation}

In other words, our goal is to generate language corrections $u$ that result in the largest decrease in discrepancy between the student and the expert. In practice, however, optimizing directly for the above expression is intractable due to the lack of strong cognitive models of human learning, i.e., we do not have an accurate model of how $u_k$ leads to changes in the student trajectory $\tau^{g,k+1}_{\mathcal{S}}$.
Therefore, instead of optimizing for the objective in Eq.~\eqref{eq:loss}, we consider whether it is possible to build a strong generative model in a supervised manner from annotated samples of corrective feedback $(\tau_\mathcal{S}^{{g}}, \tau_\mathcal{E}^{{g}}, u)$. In order to best capture the expressiveness of annotations provided in natural language, we propose leveraging the rich encoding of language present in modern day LMs by casting the problem of generating corrective feedback for student $\mathcal{S}$ in reference to  $\mathcal{E}$ as a \emph{controllable text generation} problem. Concretely, our goal is to identify a method that, given tuples of $(\tau_\mathcal{S}^{{g}}, \tau_\mathcal{E}^{{g}}, u)$, allows us to effectively control (via prompting) a large pretrained LM to generate corrections $u$ at test time when we only have access to novel student and expert trajectories $(\tau_\mathcal{S}^{{g}}, \tau_\mathcal{E}^{{g}})$.

\subsection{Trajectory Encoding}
To use trajectory samples $(\tau_\mathcal{S}^{{g}}, \tau_\mathcal{E}^{{g}})$ to construct an input prompt that can help steer an LM to generate good corrections $u$, we first need the ability to represent trajectories of a physical task as a sequence of text tokens. Recall that a trajectory $\tau$ is a sequence of state and action pairs $\{s_1,a_1, \dots, s_T,a_T\}$ which, when concatenated can be represented as a set of $T$ vectors of numerical values with dimension $d_g := [S]+[A]$.  Meanwhile, a typical LM ($\mathcal{M}_{\text{lang},\phi}$) consists of a word embedding layer ($\mathcal{W}_{\text{lang},\phi}$) that maps text tokens from a fixed vocabulary to embeddings of a given dimension $d_e$. We therefore learn a trajectory encoder model $\mathcal{M}_{\text{traj}, \theta}$ that can map any ($T\times d_g$)-dimension trajectory $\tau^g$ to a set of $n$ vectors of dimension $d_e$, where $n$ is a hyperparameter. We can then represent $\tau_\mathcal{E}^g$ and $\tau_\mathcal{S}^g$ as a sequence of ``token embeddings'' $v_{\mathcal{S},1} ... v_{\mathcal{S},n}$, $v_{\mathcal{E},1} ... v_{\mathcal{E},n}$ that,  as shown in Figure \ref{fig:model-overview}, form the input prompt to the LM which we will use to conditionally generate correction $u$.

\subsection{Controllable Text Generation}
\label{sec:traj_rep}
\model{} consists of a trainable encoder $\mathcal{M}_{\text{traj}, \theta}$ that learns to represent any arbitrary trajectory $\tau$ as a sequence of continuous embeddings such that, when embeddings corresponding to both the student and expert trajectories are included as part of a prompt, the underlying \textit{frozen}, pre-trained LM ($\mathcal{M}_{\text{lang}, \phi}$) will generate appropriate corrections. We choose to keep the LM frozen in order to aid the adaptability of \model{} to new kinds of student behavior and domains where there may be changes in language not captured by our data. 

We learn the same trajectory encoder ($\mathcal{M}_{\text{traj}, \theta}$), consisting of a 3-layer feed-forward neural network that outputs $n$ vectors with the same dimension as the target LM (e.g. 768 for GPT-2), for both student $\mathcal{S}$ and expert $\mathcal{E}$ trajectories. We train our model over tuples of corrections paired with student and expert trajectories ($\tau_\mathcal{S}$,  $\tau_\mathcal{E}$, $u)_i$ by constructing input prompt sequences using $\mathcal{M}_{\text{traj}, \theta}$ as shown in Figure \ref{fig:model-overview}. During training, we calculate the language modeling loss, where the loss of single sample $q_i$ is:
\begin{center}
    $\mathcal{L_\phi}(q_i) =  - \sum_{t=1}^{|q_i|}\text{log}\mathcal{M}_{\text{lang},\phi}(q_{i_t}|q_{i_{<t}})$
\end{center}

However, we only use  $\mathcal{L_\phi}(q_i)$ to update weights $\theta$ of the trajectory encoder $\mathcal{M}_{\text{traj}, \theta}$, keeping the weights of $\mathcal{M}_{\text{lang}, \phi}$ frozen. At test time, we use the same format (omitting $u$ which is unknown) to construct the input prompt provided to the frozen LM from which we generate corrections.  

\begin{algorithm}[tb]
   \caption{Train \model{}}
   \label{alg:train}
\begin{algorithmic}[1]
   \STATE {\bfseries Input:} dataset $\mathcal{D}$ of $(u, \tau^g_\mathcal{S}, \tau^g_\mathcal{E})$ tuples with size $|\mathcal{D}|$
   
   \STATE {\bfseries Input:}  frozen LM $\mathcal{M}_{\text{lang}, \phi}$  with token embedding layer $\mathcal{W}_{\text{lang}, \phi}$ and instruction-tuned LM $\mathcal{M'}_{\text{lang}, \psi}$  
   \STATE {\bfseries Input:} number of epochs $n_e$ ,  learning rate $\lambda$
   
   \STATE Initialize trajectory encoder $\mathcal{M}_{\text{traj}, \theta}$
   \STATE \texttt{// data augmentation}
   \STATE Set dataset $\mathcal{D'} \leftarrow \mathcal{D}$
   \FOR{sample $i=1$ {\bfseries to} $|\mathcal{D'}|$}
   \STATE Set prompt $p_i \leftarrow \textit{``You are a teacher providing''} + \textit{``feedback to a student learning a control task.''} + \textit{``List 3 short paraphrases of the feedback''} + u_i$
   \STATE Set paraphrases $u'_{i,1}, u'_{i,2}, u'_{i,3} \leftarrow \mathcal{M'}_{\text{lang}, \psi}(p_i)$
   \STATE $\mathcal{D'}$.append$((u'_{i,1}, \tau^g_{\mathcal{S}_i}, \tau^g_{\mathcal{E}_i}))$ 
   \STATE $\mathcal{D'}$.append$((u'_{i,2}, \tau^g_{\mathcal{S}_i}, \tau^g_{\mathcal{E}_i}))$ 
   \STATE $\mathcal{D'}$.append$((u'_{i,3}, \tau^g_{\mathcal{S}_i}, \tau^g_{\mathcal{E}_i}))$ 
   \ENDFOR
   \STATE \texttt{// training}
   \FOR{epoch $m=1$ {\bfseries to} $n_e$}
   \STATE Shuffle dataset $\mathcal{D'}$
   \FOR{sample $i=1$ {\bfseries to} $|\mathcal{D'}|$}
   \STATE  Set prompt $q_i \leftarrow \mathcal{W}_{\text{lang}, \phi} (\textit{student}) + M_{traj,\theta}(\tau^g_{\mathcal{S}_i}) + \mathcal{W}_{\text{lang}, \phi}(\textit{expert}) + M_{traj,\theta}(\tau^g_{\mathcal{E}_i}) + \mathcal{W}_{\text{lang}, \phi} (\textit{correction:} )+\mathcal{W}_{\text{lang}, \phi}(u_i) $
   \STATE Set loss $\mathcal{L}(u_i, \tau^g_{\mathcal{S}_i}, \tau^g_{\mathcal{E}_i}) \leftarrow \mathcal{L}_\phi(q_i)$  \texttt{LM loss}
   \STATE Update $\theta \leftarrow \theta + \lambda \nabla_{\theta}\mathcal{L}(u_i, \tau^g_{\mathcal{S}_i}, \tau^g_{\mathcal{E}_i})$
   \ENDFOR
   \ENDFOR
\end{algorithmic}
\end{algorithm}

%Megha: add a sentence saying that it is challeing to find such data elsewhere due to the expoential space? 
\subsection{Annotating Corrections \& Data Augmentation}
In order to train \model{}, we need to collect data of corrections for paired trajectories. Because our goal is for \model{} to generalize well to novel trajectories and domains, we are primarily interested in shorter, general corrections that do not refer to specific aspects of the expert's trajectory or domain-specific objects. 
Concretely, we ask annotators to provide brief samples of corrective feedback $u^{(1)}, u^{(2)}, ..., u^{(m)} $ for a particular $\tau^g_\mathcal{S}, \tau^g_\mathcal{E}$ trajectory pair for task $g$ in free-form text, encouraging annotators to identify which of the potentially several different ways for the student to improve they believe is most optimal to describe. We can then use tuples $(\tau^g_\mathcal{S}, \tau^g_\mathcal{E}, u^{(i)})$ to construct input prompts to train \model{}. Further details on crowdsourcing results of for our annotation procedure are described in Section \ref{sec:crowdsourcing} . 

However, we observe that when human annotators provide corrective feedback in natural language, there exists greater variance in the language style of the provided corrections than the particular discrepancies they refer to.  In order to enable \model{} to better capture this rich style diversity efficiently,  we leverage more powerful, ``instruction tuned'' language models (e.g. OpenAI's \texttt{text-davinci-003}) for data augmentation. As described in Algorithm \ref{alg:train}, for each annotation $u^{(i)}$ in our original dataset, we construct an input prompt describing a teaching setting and directly asking for paraphrases of $u^{(i)}$, which, when sent as input to a large instruction-tuned LM results in an augmented set of utterances $\{u'^{(i)}_1, u'^{(i)}_2, u'^{(i)}_3\}$ which are used for training. The prompt and example paraphrases are shown below:

\begin{center}
    \small{\textbf{annotator correction:} \\ turner slightly later $(u)$} \\ 
    \small{\textbf{input prompt:} \\ You are a teacher providing feedback to a student learning a control task. List 3 short paraphrases of the feedback \textit{``turner slightly later"}} \\ 
     \small{\textbf{text-davinci-003 output:} 
     \\ 1. Make your turn a bit later. $(u'_1)$ \\ 2. Delay your turn a bit $(u'_2)$ \\ 3. Wait a moment before turning $(u'_3)$} \\
\end{center}

The above example shows that paraphrases returned from the \texttt{text-davinci-003} LM retain the particular discrepancy of the correction while modifying its style, language, and correcting for typos and grammatical errors. 
As we will show next (Table \ref{table-ppl}), training \model{} over augmented data improves performance across all control tasks.

\begin{figure*}
\vskip 0.2in
\begin{center}
\centerline{\includegraphics[width=\linewidth]{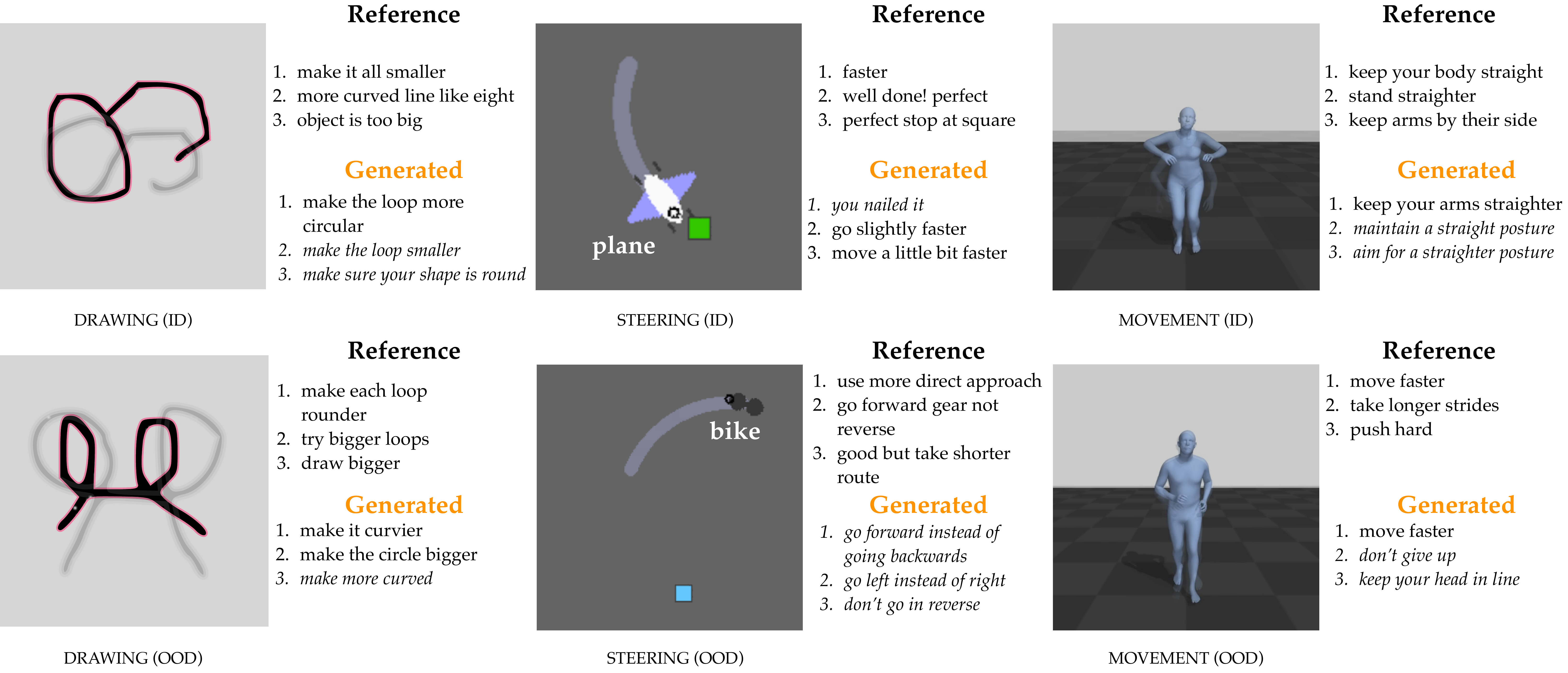}}
\caption{Example student trajectories, reference corrections from annotators, and corrections generated by \model{} for novel trajectories for all three control tasks. Generated corrections in \textit{italics} are completely unseen during training, for any trajectory.}
\label{fig:experiment-overview}
\end{center}
\vskip -0.2in
\end{figure*}

\begin{table*}[t]
\caption{Perplexity on held-out test sets (lower is better) across three control tasks. \model{} achieves lower perplexity in comparison to baselines across all tasks, and both pre-training and data augmentation components improve performance. Although there exists a gap between  in-domain (ID) and out-of-domain (OOD) performance, 
\model{} still outperforms ablations even in OOD settings. }
\label{table-ppl}
\vskip 0.15in
\begin{center}
\begin{small}
\begin{sc}
\begin{tabular}{lcccccccr}
\toprule
\multicolumn{1}{c}{Ablation} & \multicolumn{2}{c}{Drawing} & \multicolumn{2}{c}{Steering} & \multicolumn{2}{c}{Movement} \\
\midrule
 & ID & OOD & ID & OOD & ID & OOD \\
\midrule
Permute Correction & 310 $\pm$ 38 & 249 $\pm$ 1.1 & 84 $\pm$ 18.5 & \textbf{194 $\pm$ 2.4} & 47 $\pm$ 2.3& 123 $\pm$ 7.4\\
Permute Student   & 153 $\pm$ 5.6 & 256 $\pm$ 5.9 & 96 $\pm$ 8.9& 218 $\pm$ 3.1 & 35 $\pm$ 0.28 & 111 $\pm$ 4.9\\
\midrule
\model    & \textbf{145 $\pm$ 1.5} &  \textbf{246 $\pm$ 2.5} &  \textbf{51 $\pm$ 5.9}  & \textbf{194 $\pm$ 2.3} & \textbf{33 $\pm$ 0.22} & \textbf{109 $\pm$ 3.1}        \\
w/o Data Aug.     & 162 $\pm$ 6.3  &  251 $\pm$ 2.9 & 54 $\pm$ 1.8& 635 $\pm$ 24.3 & 36 $\pm$ 2.3 & 159 $\pm$ 6.7\\
w/o Pretraining (GPT-2) & 959 $\pm$ 62 & 808 $\pm$ 72 & 302 $\pm$ 32 & 848 $\pm$ 88& 376 $\pm$ 37 & 823 $\pm$ 53\\
w/o Pretraining (LSTM) & 215 $\pm$ 1.2 & 584 $\pm$ 1.2 & 197 $\pm$ 1.4 & 271 $\pm$ 1.1& 221 $\pm$ 1.3 & 252 $\pm$ 1.1\\
\bottomrule
\end{tabular}
\end{sc}
\end{small}
\end{center}
\vskip -0.1in
\end{table*}

\begin{table*}[t]
\caption{Similarity scores on held-out test sets (higher is better) based on an improved BERTScore to account for ground truth variance from \citep{yi2020captioning}. Across all tasks, \model{} outperforms both randomly sampling ID feedback and a nearest neighbors baselines.}
\label{table-similarity}
\vskip 0.15in
\begin{center}
\begin{small}
\begin{sc}
\begin{tabular}{lcccccr}
\toprule
\multicolumn{1}{c}{Method} & \multicolumn{2}{c}{Drawing} & \multicolumn{2}{c}{Steering} & \multicolumn{2}{c}{Movement} \\
\midrule
 & ID & OOD & ID & OOD & ID & OOD \\
\midrule
 Random  & 0.20 $\pm$ 0.03 & 0.21 $\pm$ 0.04 & 0.19 $\pm$ 0.04 & 0.22 $\pm$ 0.03 & 0.23 $\pm$ 0.06 &   0.18 $\pm$ 0.03       \\
 Nearest Neighbors  & 0.28 $\pm$ 0.03 & 0.22 $\pm$ 0.03 & 0.28 $\pm$ 0.05 &  0.16 $\pm$ 0.04 &  0.31 $\pm$ 0.05 & 0.19 $\pm$ 0.05    \\
 Permute Student  & 0.22 $\pm$ 0.03 & 0.23 $\pm$ 0.04 & 0.14 $\pm$ 0.03 &  0.26 $\pm$ 0.01 &  0.14 $\pm$ 0.03 & 0.15 $\pm$ 0.03   \\
\midrule
\model    & 0.3 $\pm$ 0.01 & \textbf{0.34 $\pm$ 0.03} &   \textbf{0.32 $\pm$ 0.08} & \textbf{0.31 $\pm$ 0.02} & \textbf{0.39 $\pm$ 0.03} & \textbf{0.24 $\pm$ 0.03}      \\
w/o Pretraining (GPT-2) & 0.11 $\pm$ 0.02 & 0.18 $\pm$ 0.03 & 0.10 $\pm$ 0.03 & 0.12 $\pm$ 0.03 & 0.11 $\pm$ 0.03 & 0.11 $\pm$ 0.02 \\
w/o Pretraining (LSTM) & 0.15 $\pm$ 0.03 & 0.17 $\pm$ 0.03 & 0.12 $\pm$ 0.04 & 0.13 $\pm$ 0.03 & 0.15 $\pm$ 0.03 & 0.18 $\pm$ 0.02 \\
w/o Data Aug.     & \textbf{0.32 $\pm$ 0.04} &  0.26 $\pm$ 0.04 &  0.26 $\pm$ 0.03 & 0.27 $\pm$ 0.02  & 0.19 $\pm$ 0.05  & 0.23 $\pm$ 0.02 \\
\bottomrule
\end{tabular}
\end{sc}
\end{small}
\end{center}
\vskip -0.1in
\end{table*}

\section{Experimental Results}
We now present our three tasks and  experimental results. Details of user studies (including IRB approval) and training of \model{}, which is built on a 124M parameter model of the GPT-2 family \citep{huggingface}, are in the Appendix. 
\subsection{Environments \& Datasets}
We study three physical control tasks that span common primitives: drawing (x-y control), steering (acceleration and heading angle control), and human body movement (joint control). For each environment, we also create in-domain (ID) and an out-of-domain (OOD) splits that share the same control space, but require different dynamics.\footnote{While we aimed to pick OOD splits that were semantically far (e.g. Futurama is a synthetic language), it is still possible there may be smaller ``sub-skills'' shared between ID-OOD splits.}

    \textsc{Drawing}: The student's goal is to learn how to draw characters from different alphabet scripts.  We select 10 characters from 5 scripts (ID: Arabic, Burmese, \& Japanese, OOD: Futurama \&  Bengali) from the Omniglot dataset \citep{lake2015human}. We select 1 trajectory per character as the expert trajectory and randomly sample 5 student trajectories, split between train/test sets. Each trajectory is a sequence of 2D actions along x-y coordinates.  
    
    \textsc{Steering}: The student's goal is to learn how to park a vehicle in a target parking spot. We modify the Parking environment from \citet{highway-env} by changing the steering sensitivity and min/max speed for 3 vehicle types  (ID: Car \&  Plane, OOD: Bike). For each vehicle type, we design a hand-coded expert policy, and then collect 20 student trajectories including perturbations of the expert policy and half-trained RL agents (details in  Appendix \ref{sec:training}). Trajectories are split between train/test sets, and consist of 2D actions controlling acceleration and heading angle and 6D states corresponding to vehicle position, velocity, and heading.  
    
    \textsc{Movement}: The student's goal is to learn how to perform a full-body movement activity. We select activities from the BABEL dataset \citep{babel} of 3D human motion (ID: Walk, Jump, \& Throw, OOD: Wave, Jumping Jacks). For each activity we select 1 trajectory as the Expert, and sample 15 student trajectories, which are then split between train/test sets. We represent trajectories with learned video-text representations from X-CLIP \citep{ni2022expanding}, treating the output as a trajectory sequence of 1D states.   

Example student trajectories for each environment are shown in Figure \ref{fig:experiment-overview}. We pad trajectories to a fixed dimension of 10 and length of 600 as input to \model{}. Further details on expert trajectory selection, as well as the assumption of a single expert behavior, are in Appendix \ref{sec:single}. 

\subsection{Crowdsourcing Details}
\label{sec:crowdsourcing}
We recruit crowdworkers on  Prolific\footnote{https://www.prolific.co/} to annotate paired student/expert trajectories with corrections. We instruct crowdworkers to not refer to expert demonstrations in their annotations. Crowdsourced corrections demonstrate a variety of ways people express feedback, such as rich shape descriptions (e.g. \textit{``go towards making an infinity shape rather than a venn diagram''}), encouragement (e.g. \textit{``more vertical but good effort'' }), and action ordering (e.g. \textit{``after second bend draw towards left not down''}. We collect \textbf{2,023} corrections, and provide further details in Appendix \ref{sec:crowdsourcing-app}.

\subsection{Automatic Evaluation}
\label{sec:auto-eval}
%education motivation
Our first evaluation goal is to measure the degree \model{} assigns high likelihood to examples of good corrections, which can be useful for tasks such as automatically evaluating feedback provided by instructors. In Table \ref{table-ppl}, we report the average perplexity (i.e. the exponentiated loss) across ground truth corrections for novel student trajectories unseen during training, and for both ID and OOD splits of each task. We compare results across the following ablations: 

\begin{itemize}
    \item \textbf{Permute Correction}: Instead of conditionally generating a correction, we draw a random corrections from the same distribution as the ground-truth corrections -- if a task has low variance across the types of feedback needed (e.g. all students need to ``improve posture'' in \textsc{movement}), we should observe no difference. 
    \item \textbf{Permute Student}: We simulate the setting where \model{} provides corrective feedback for a different student trajectory. This measures the degree \model{} may have only fitted to the fixed expert trajectory -- it  should assign higher (worse) perplexity when the student trajectory is randomized, showing the ability to tailor corrective feedback to individual students. For fair comparison, we  sample student trajectories from the eval set to maintain the same overall distribution. %not expert specific
    \item \model{} \textbf{w/o Pretraining }: We ablate the effect of pretraining by (i) using the same GPT-2 architecture, but without pre-trained weights, and (ii) using a 3-layer LSTM with pre-trained embedding layer.
    \item \model{} \textbf{w/o Data Augmentation}: We train \model{} on the original, smaller dataset consisting purely of human annotations, without any paraphrases from our automatic data augmentation procedure. 
\end{itemize}

As Table \ref{table-ppl} shows, \model{} outperforms both permutation ablations, suggesting that the model does take into account specific student trajectories, rather than just learning general task language. 
As expected, no pre-training decreases performance, due to the lack of strong language representations. Furthermore, data augmentation results in an improvement across all tasks for both ID and OOD settings. Although the gap between ID and OOD is high, we note that even in OOD settings \model{} generally outperforms ablations.
%, and that this may likely be due to perplexity penalizing for the OOD trajectory types, when we are primarily interested in the quality of the correct feedback utterances. 

Thus, our second automatic evaluation focuses on the quality of generated samples from \model{}. Under a fixed set of decoding parameters (nucleus sampling \citep{holtzman2020curious}, temperature = 0.5), we measure the average similarity between generated and ground-truth corrections across each ($\tau_\mathcal{S}$,  $\tau_\mathcal{E})_i$ in our test set.  However, as Figure \ref{fig:experiment-overview} shows, annotations for a sample may have high variance due to identifying different discrepencies. We therefore use a re-weighted version of BERTScore that accounts for intrinsic variance between ground-truth captions, originally proposed for image captioning \citep{yi2020captioning}. In addition to the pre-training and data augmentation ablations, we compare the average similarity across generated samples from three alternative methods with \model{}:

\begin{itemize}
    \item \textbf{Random:} We select a random human annotation from the same domain as the input trajectories, allowing us to measure the degree \model{}'s performance is due to just using vocabulary appropriate for the domain. 
    \item \textbf{Nearest Neighbors:} For a given student trajectory in our test data, we use our trajectory encoder $\mathcal{M}_{\text{traj}, \theta}$ to find the nearest neighbor student trajectory seen during training (using the mean squared error in encoder output). We then randomly sample from the set of ground-truth annotations provided for this student. 
    \item \textbf{Permute Student:} We select a correction from the same domain and expert as the input trajectories, but a random student. Note this method is distinct from the Permute Student method in the previous section. 
\end{itemize}

Table \ref{table-similarity} shows that \model{} outperforms both methods across all tasks, for both ID and OOD settings. As expected, removing pre-training results in samples with lower similarity scores than \textbf{Random}, and we observe that without using a pre-trained LM, the model can only generate domain specific verbs (e.g. \textit{``make''} or \textit{``move''}). Interestingly, we observe that for this metric, there is less of a gap between ID and OOD -- in fact, for \textsc{Drawing}, generated samples from \model{} are \textit{more} similar to ground truth annotations for OOD characters. As shown in Figure \ref{fig:experiment-overview}, for both ID and OOD  we observe that \model{} indeed often generates corrections that are similar to the ground-truth annotations. 

\textbf{Error Analysis} \\
In practice, however, neither automatic evaluation metric we report fully captures the complexities of evaluating corrections. For example, the types of sequences \model{} assigns high (worse) perplexity to include metaphorical utterances and noise (e.g. \textit{``the shape at the top should be larger, marching the hook shape''}) and domain-specific language (e.g. \textit{``go forward gear not reverse''}).  Meanwhile, the improved BERTScore method from \citet{yi2020captioning} assigns a score of 0.0 to examples such as \textit{(\textbf{reference}: well done, perfect!, \model{} : you nailed it!)}, where the expressed meanings are equivalent, but use very different language. This motivates the need for human evaluation, which we focus on next.   

\subsection{Human Preference Evaluation}
\begin{figure*}[t]
\vskip -0.1in
\begin{center}
\centerline{\includegraphics[width=\linewidth]{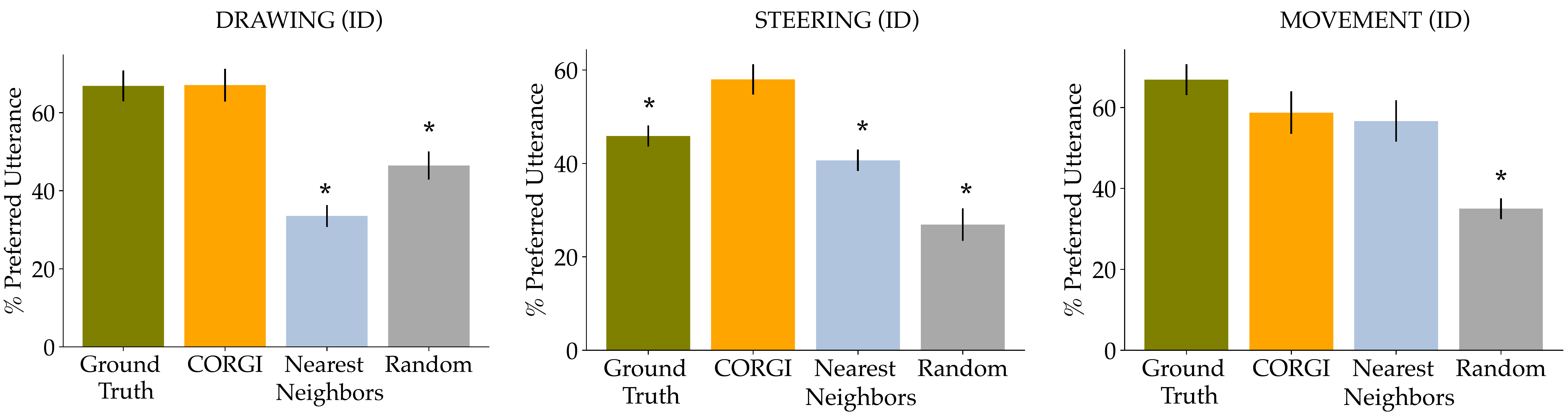}}
\caption{Across all three tasks, users are more likely to prefer feedback generated from \model{} over random corrections than feedback from a random control and nearest neighbors baseline. For \textsc{steering}, feedback from \model{} also outperforms ground truth corrections, which may be due to the high variance human annotations.  Asterisk (*) marks statistically significant difference ($p < 0.05$) from  \model{} .}
\label{fig:user_preference}
\end{center}
\vskip -0.4in
\end{figure*}

 We first choose to assess the degree human evaluators \textit{prefer} \model{} over randomly chosen utterances from the same domain. Specifically, we measure preference as the rate at which human evaluators prefer the correction that is generated by \model{} when provided two other randomly selected corrections from the same domain. We then compare this rate with three other conditions that replace \model{}: 

\begin{itemize}
    \item \textbf{Random:}  We calculate the rate at which human evaluators pick a correction randomly selected from the training data within the same domain. Since the other options are also randomly sampled, as the number of samples increase, this should converge to $33\%$. 
    \item \textbf{Nearest Neighbors:} Already described in section \ref{sec:auto-eval}, we randomly sample a ground-truth correction provided to the nearest neighbor student. 
    \item \textbf{Ground Truth:}  We calculate the rate at which human evaluators pick a corrections sampled from the set of ground-truth annotations for the target trajectory. 
\end{itemize}

Users are shown a pair of student and expert trajectories (e.g. videos of human movement for \textsc{movement}) and asked to pick one of the three corrections in response to the instruction \textit{``Which feedback do you think is most helpful to provide to the student?''}.
 We collect preference data from 15 users  per condition for each of our three tasks, randomizing the order in which each correction is provided. We recruit crowdworkers on Prolific, and provide further details in Appendix \ref{app:human-pref}. Due to cost, we limit ourselves to only novel  in-domain (ID) trajectories for each of our control tasks. 

Figure \ref{fig:user_preference} shows that across all three control tasks, users were significantly more likely to prefer corrections from \model{} than our \textbf{Random} control. Furthermore, corrections generated with the \textbf{Nearest Neighbors} method are only comparable to those of \model{} for the \textsc{movement} task, highlighting the ability of \model{} to generalize to student trajectories unseen during training. Surprisingly, in the \textsc{steering} task, we observe that \model{} significantly outperforms \textbf{Ground Truth}. One potential hypothesis is that preferences capture important aspects of corrections beyond accuracy, including clarity, constructiveness, and tone. Generated samples from \model{} are often concise and formal, while human corrections exhibit more variety. For example, the most common human annotation that evaluators did \textit{not} select in the \textsc{steering} task was \textit{``right hand down, route south''}, which may be less clear than the generated sample for the same comparison (\textit{``glide gracefully to the left''}).  Finally, we provide pair-wise comparison results on feedback from \model{} when directly compared with \textbf{Ground Truth} and \textbf{Nearest Neighbors} feedback in Appendix \ref{app:human-pref}.

\subsection{Learning from Feedback}
Our final human evaluation directly measures the degree \model{} helps reduce the discrepancy between student $\mathcal{S}$ and expert $\mathcal{E}$ performance in the \textsc{drawing} task. We design a teaching interface, shown in Appendix \ref{app:human-learn}, where users are given three chances to draw a provided stimulus and match a hidden expert trajectory $\tau_\mathcal{E}$. The only information users receive are  corrections corresponding to their trajectory $\tau_\mathcal{S}$, and a numerical score calculated with the mean squared error between $\tau_\mathcal{S}$ and expert trajectory $\tau_\mathcal{E}$. We then measure the change in student error between the first and third trial. 

We assign 20 users to a control group where corrections are randomly sampled from data within the same domain, 20 users to a control group where no corrections are provided, and  20 users to the experiment group, who receive corrective feedback from \model{}. While  users who received random feedback (\textbf{-0.17 $\pm$ 1.16}) and no feedback (\textbf{-0.20 $\pm$ 1.01}) both on average \textit{decreased} in performance, users provided feedback from \model{} actually \textit{improved} with an average score difference of  \textbf{1.84 $\pm$ 0.7}. 
A larger sample size may be needed to observe a stronger effect (we observe $p < 0.1$ using a Welch's t-test with multiple hypothesis correction, verifying normality assumption and medium effect size of Cohen's $d=0.52$). However, we provide further results showing that feedback from \model{} also outperforms a baseline with only visual feedback, and covers a diverse set of topics such as size (\textit{``make it all a bit bigger''}) and edge straightness,  in the Appendix.

Overall, our results show that \model{} can generate corrective feedback for novel student trajectories across a diverse set of control tasks that not only outperform baselines in automatic evaluation, but are also preferred by human raters and help learners improve at a physical control task. One appealing aspect of \model{} is the ability to avoid fine-tuning the underlying LM. This allows us to retrain the rich and expressive encoding the LM has learned, enabling several possible directions for future work that we discuss next.  
%These larger models also exhibit the emergent behavior of in-context learning \megha{cite}, where users simply provide text examples of a new task they wish the model to perform in the input prompt. Looking forward, we might expect similar behavior to arise in large, multi-modal models that can handle trajectory inputs -- it is therefore natural to try to understand the degree models trained on large scale internet text data can encode different physical concepts required for control tasks, which we explore next.

% Looking Forward: Zero-Shot Capabilities
% Recent work, as part of a larger debate on whether large-scale language models to encode ``meaning'', have shown that such models can encode physical conceptual spaces, such as cardinal directions, as well as implicit models of meaning that can be used to describe dynamic state representations \citep{li2021implicit, patel2022conceptual}. In this exploratory analysis, we expand upon the method proposed by \citet{patel2022conceptual} to study comparative properties more relevant to physical control tasks: \textit{angle, force, speed, and height}. Using a synthetic grid world set up, as shown in \megha{REF}, we construct prompts consisting of sequences of student and expert trajectories, and measure the zero-shot generalization capabilities of several large-scale language models on novel ``worlds''. Specifically, we \megha{TODO, measure accuracy} 
% \megha{as LM size increases, keep the world the same allowing comparitive understanding of language}

%height width dimension

\section{Limitations \& Future Directions}
As our work is a first step towards building a model capable of generating natural language corrections for physical control tasks, there are a few limitations and important directions for future work.
First, one important aspect of corrective feedback is \textit{tone}: language with positive encouragement may lead to different student learning outcomes than more terse feedback, and future work could consider adding information about the student (e.g. age, personality) as an additional control for \model{}. 

Another limitation is that \model{} does not generate feedback with domain-specific references -- future work could consider integration of corrections from \model{} with domain-specific approaches \citep{schrum2022reciprocal}. Additionally, while \model{} only provides corrections over the entire trajectory, many control tasks involve complex sequences of actions that combine many different sub-tasks, or skills. Future work could consider learning how to jointly break down student trajectories into different sub-components, and then generating corresponding feedback for each part. 

Finally, as described in Appendix \ref{sec:single}, a key assumption of our work is the need for an expert reference trajectory used to provide feedback. In practice, there may be many expert ways to perform a physical control task, which expert-specific systems may fail to capture. While \model{} can flexibly take any expert trajectory as input, its performance is limited by the diversity of expert trajectories it saw during training, and we believe enabling \model{} to generate appropriate corrections for a diverse range of expert behaviors in a data efficient manner is an important next step.

Finally, because \model{}  can take any student and expert trajectory as input, potential misuse includes a malicious agent leveraging \model{}  repeatedly to generate corrections that actually guide a student towards harmful behavior (e.g. physical actions that harm the body). An interesting avenue for future work is creating a mechanism that can detect whether an expert trajectory is plausible and safe for a human to perform under domain-specific constraints.

%Finally, in addition to our dataset of 2k corrective feedback annotations, data from our human preference experiment sheds light on the nuances of how humans interpret and receive different forms of feedback. Complementing existing work on RL from human feedback \citep{christiano2017deep}, it is exciting to consider whether such preference data can help us improve \model{} over time and adapt to the way people may change how they provide corrective feedback.
%RLHF from human preferences, incorporate 
\section{Acknowledgements}
We thank all reviewers for their valuable feedback. We acknowledge support from Point72, Ford, AFOSR, and NSF Awards \#2218760, \#2132847, and \#2006388. MS was also supported by the NSF GRFP under DGE-1656518.

% Note use of \abovespace and \belowspace to get reasonable spacing
% above and below tabular lines.

\bibliography{main}
\bibliographystyle{icml2023}

\newpage
\appendix
\onecolumn
\section{Appendix}
\label{appendix:all}
We include information about accessing our dataset, model checkpoints, and user study infrastructure at this link: \url{https://github.com/Stanford-ILIAD/corgi}.

\subsection{Ethics Statement \& IRB}
The purpose of our work is to help student learners improve performance on control tasks by automatically generating fluent and accuracy feedback in natural language. However, because physical control tasks can affect user comfort and health, an important risk of our work is its potential to mislead a person to perform control movements that may be harmful. Furthermore, a malicious actor can leverage the method behind \model{} to train a model that intentionally hurts user performance. For these reasons, we emphasize the importance of ensuring safety checks when deploying a system based on \model{} and exercising caution in critical application areas.

Human subject studies, including both the human preference and learning performance evaluations, were conducted as part of a study approved by Stanford University's Institutional Review Board (protocol \# IRB-49406). Participants were asked to agree to a consent form (like \href{https://drive.google.com/file/d/1pOeYx5Bd-qjlCKmD2fJUzome4xmZbQYW/view}{this example}), before continuing to the study interface. All participants were crowdworkers recruited on the Prolific platform. 

\subsection{Expert Trajectory Assumptions}
\label{sec:single}
One important assumption of our work is the need for an expert reference trajectory used to provide feedback. While all experiments in this work are conducted with a limited range of experts, in reality there exist multiple expert behaviors for a task (e.g. using the right hand or the left hand) that result in different trajectories. An ideal teaching system would be able to take as input any arbitrary expert behavior, and provide appropriate corrective feedback for the system. While \model{} has this capability with respect to its API (any arbitrary expert behavior can be sent as input), we chose not to cover an exhaustive range of expert behavior due to nuance in defining different ``optimal'' experts: for example, in the \textsc{drawing} task, while drawing the letter ``I'' bottom-up or top-down might be equally optimal, this may not be true for particular applications like rehabilitation, where a trained may seek to guide a student towards a specific expert behavior. Furthermore, we believe one important aspect of good teaching is developing strong priors on the types of mistakes a student might make for a given task. For example, before even observing a student, a tennis instructor may know that hitting a ball too low is a common mistake. Training a model over a selected set of expert references, rather than across any possible trajectory as an expert, can help provide this inductive bias. Nevertheless, we introduce variance in expert trajectories for each task by (i) varying characters for \textsc{Drawing}, (ii) perturbations to expert trajectories in \textsc{Steering}, and (iii) multiple expert demonstrations for \textsc{Movement}. Future work could consider training on more varied expert behavior as well as designing a system to identify which expert behavior to provide as input to \model{}, depending on the student's learning preferences.

\subsection{Training Details}
\label{sec:training}
The trajectory encoder $\mathcal{M}_{\text{traj}, \theta}$ part of \model{} is trained for 200 epochs on one NVIDIA A40 GPU with a batch size of 64 and learning rate of 0.05, although we observed little sensitivity in performance with respect to learning rate. We split our training dataset into train and valid splits, and use the latter to perform early stopping. We repeat the same training procedure for both model ablations (no pre-training and no data-augmentation). The frozen LM we use is the 124M-parameter version of GPT-2 from \citet{wolf2019transformers}. 

We set the parameter $n$ for  $\mathcal{M}_{\text{traj}, \theta}$ to be 20, so the trajectory encoder outputs a set of 20 vectors with dimension 768. $\mathcal{M}_{\text{traj}, \theta}$ is a 3-layer feed-foward neural network, where each layer has an output size of $n=20 \times 768$.

For the \textsc{steering} task, we use trajectories from partially-trained Soft Actor-Critic agents trained for only 100 epochs using the StableBaselines3 implementation as some of our student trajectories. This leads to a variety of failure modes, which we human annotators describes. 

\subsection{Crowdsourcing Language Corrections}
\label{sec:crowdsourcing-app}
\begin{figure}[h]
\label{app-fig:crowd}
\vskip -0.1in
\begin{center}
\centerline{\includegraphics[width=0.7\linewidth]{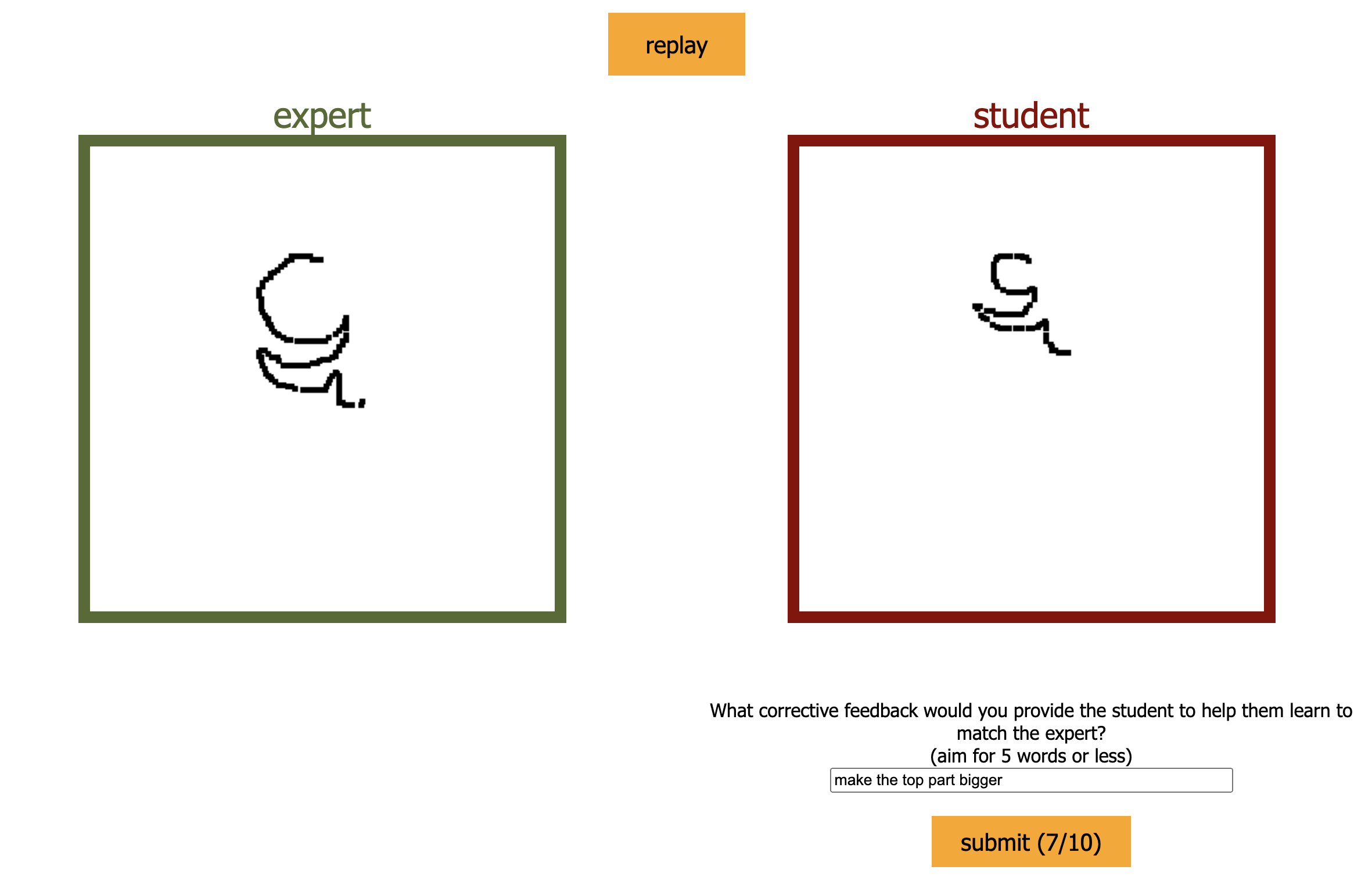}}
\caption{User Interface for Crowdsourcing Corrections for the Drawing task}
\end{center}
\vskip -0.2in
\end{figure}

For each of our three control tasks, we recruit crowdworkers on Prolific to provide corrective feedback to a student given pairwise student and expert trajectories, as seen in Figure 4. Each crowdworker provides 10 language corrections, and we pay then 14 USD per hour. In total, we collect \textbf{2,023} corrections. 
\subsection{Human Preference Evaluation}
\label{app:human-pref}
\begin{figure}[h]
\label{app-fig:pref}
\vskip -0.1in
\begin{center}
\centerline{\includegraphics[width=0.7\linewidth]{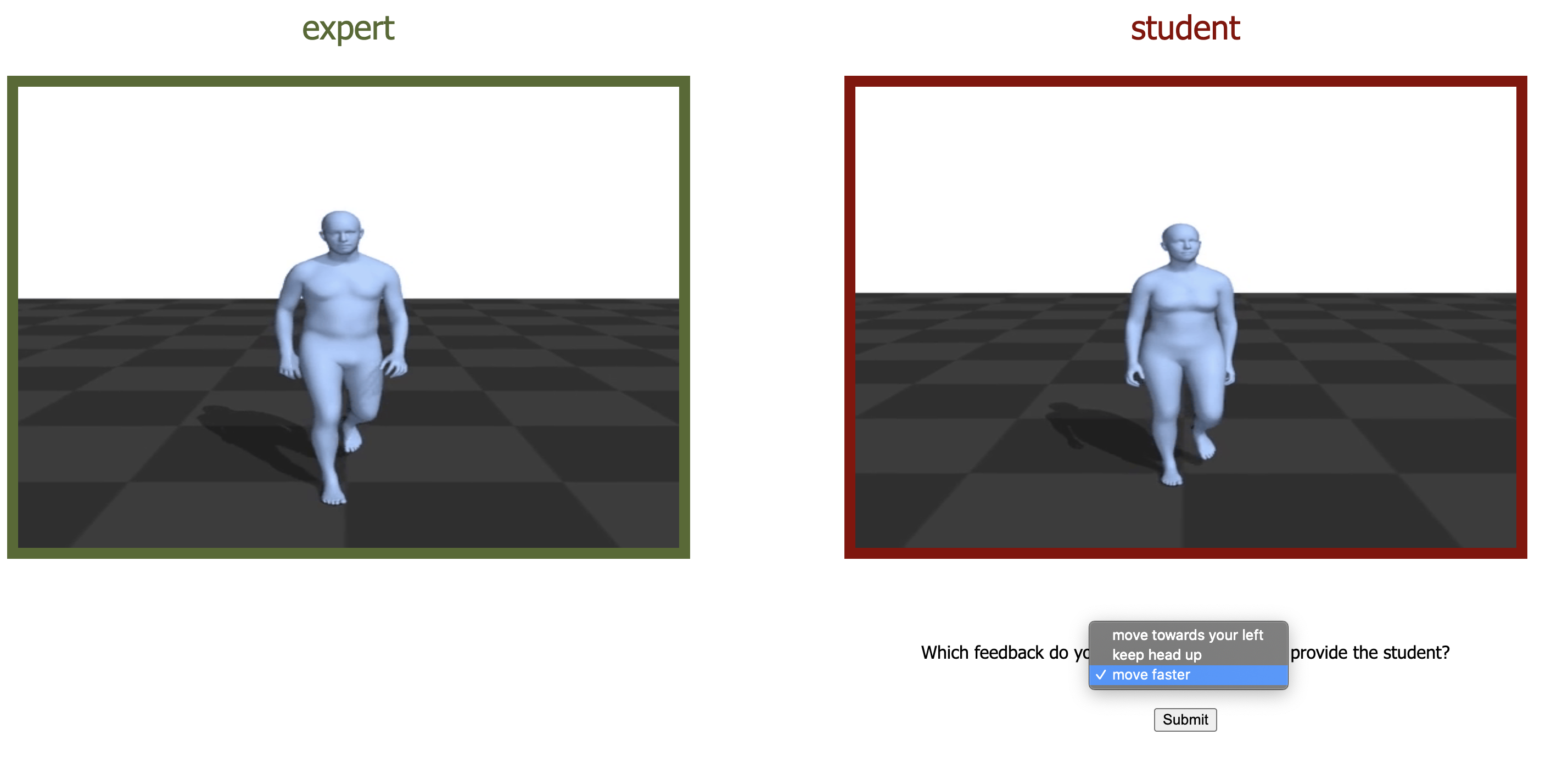}}
\caption{User Interface of the Human Preference Evaluation}
\end{center}
\vskip -0.2in
\end{figure}
For each of our three control tasks, we recruit crowdworkers on Prolific to select their preferred feedback to provide to a student given pairwise student and expert trajectories and a dropdown list of language corrections to pick from, as seen in Figure 5. Each crowdworker provides 10 preferences, and we pay then 14 USD per hour. In total, we collect \textbf{1,800} preference ratings. 

\paragraph{Direct Pairwise Comparisons}
In addition to our main results, we use the same interface to conduct direct pairwise comparisons from participant preferences between feedback from \model{} vs. \textbf{Nearest Neighbors} and feedback from \model{} vs. \textbf{Ground Truth}. We report these results in Table \ref{table-pref-direct}, which support the results reported in the main paper: \model{} outperforms the \textbf{Nearest Neighbor} baselines significantly for \textsc{Writing} and \textsc{Steering} tasks, and even outperforms \textbf{Ground Truth} annotations for the \textsc{Steering} task.

\begin{table*}[t]
\caption{Users are significantly more likely to prefer \model{} over \textbf{Nearest Neighbors} for the \textsc{Writing} and \textsc{Steering} tasks, and even outperforms Ground Truth feedback for the \textsc{Steering} task. Asterisk (*) marks results that are statistically significant ($p < 0.05$) with multiple hypothesis correction, using a binomial test where the null hypothesis is set to equal preference rate.}
\label{table-pref-direct}
\vskip 0.15in
\begin{center}
\begin{small}
\begin{sc}
\begin{tabular}{lcccr}
\toprule
\midrule
 Domain & \% CORGI Preferred 
vs. Nearest Neighbors  & \% CORGI Preferred 
vs. Ground Truth \\
\midrule
Writing & $74 \pm 4.1^*$ & $58 \pm 5.6$ \\
\midrule
Steering & $59 \pm 3.7^*$ & $60 \pm 3.3^*$ \\ 
\midrule
Movement & $54 \pm 3.1$ & $45 \pm 2.5$ \\
\midrule
\bottomrule
\end{tabular}
\end{sc}
\end{small}
\end{center}
\vskip -0.1in
\end{table*}

\subsection{Human Learning Evaluation}
\label{app:human-learn}

We evaluate the degree corrections from \model{} help humans learn for the \textsc{drawing} task by recruiting 60 crowdworkers on Prolific, split evenly between two  control groups (random feedback and no feedback) and the experiment group, to try drawing a target stimulus as seen in Figure 6. Each crowdworker provides three drawing trajectories, and we measure the difference between the third and first trial in terms of error with respect to the (hidden) expert trajectory. We pay each crowdworker 14 USD per hour. Example user trajectories can be seen in Figure 7.

Feedback generated from \model{} covers a diverse set of topics for participants in our user study. While find that $~70\%$ of corrections focus on size (split evenly between increasing and decreasing size), several participants received feedback about line sharpness (e.g. 13\%) and straightness (10\%). Additionally,  there was a long tail of corrections that were only generated once for a student (e.g. “make it stronger”, referring to the drawing line weight). Even for corrections referring to size, there exists variation in the degree of the correction (e.g. ``needs to be a bit larger'' vs. ``make it smaller''). 

\paragraph{Visual Feedback Comparison}
Finally, we run an additional experiment evaluating providing visual feedback, instead of language feedback, by providing a visual overlay on the drawing canvas. This naturally makes the task easier  for more stationary environments like drawing. However, observations from a user study conducted don 20 additional crowdworkers recruited on Prolific show that while  indeed participants perform on average around \textbf{10.1} points (between 0 and 100) higher in overall task performance than students receiving language feedback from CORGI, the learning gain (change in error from expert trajectory) is \textbf{0.39 +/- 0.48}, which is lower than those provided language feedback from CORGI. This is likely because learners, when given access to a visual overlay for this task, can immediately start to perform well, while language identifying specific areas to improve on can be remembered long-term by students. 
 
\newpage

\begin{figure}[h]
\label{app-fig:learn}
\vskip -0.1in
\begin{center}
\centerline{\includegraphics[width=0.7\linewidth]{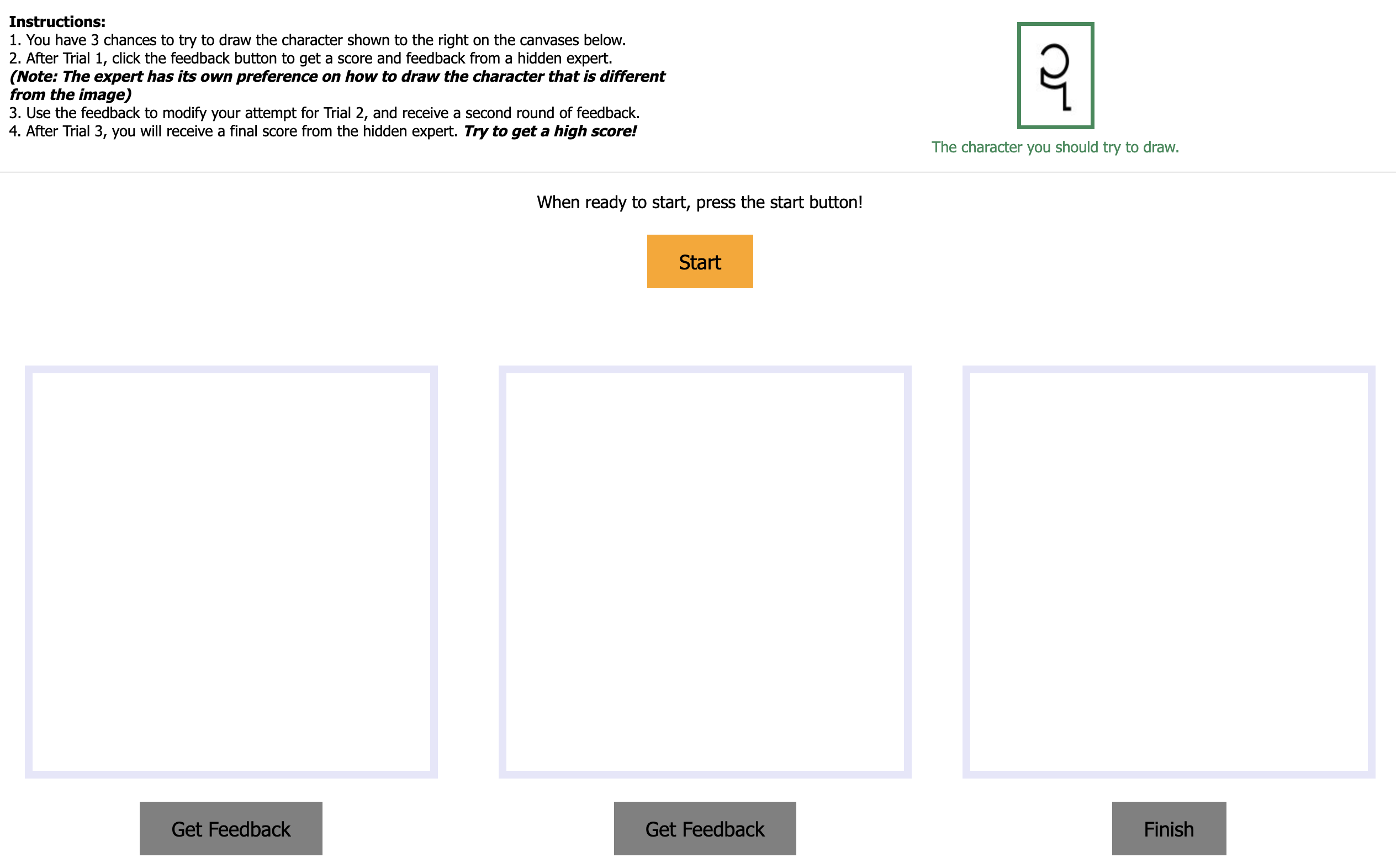}}
\caption{User Interface of the Human Learning Evaluation}
\end{center}
\vskip -0.2in
\end{figure}

\begin{figure}[h]
\label{app-fig:traj}
\vskip -0.1in
\begin{center}
\centerline{\includegraphics[width=0.7\linewidth]{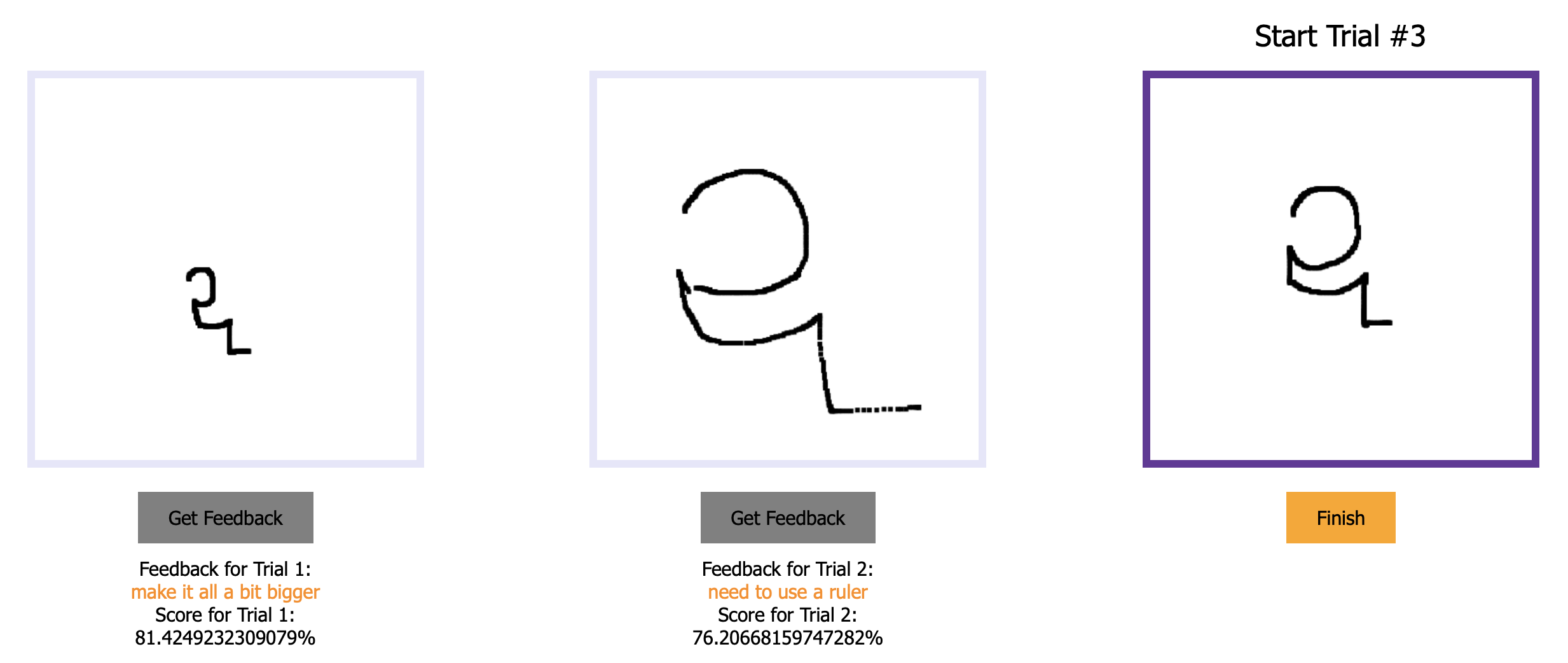}}
\caption{Example User trajectories with feedback from our model}
\end{center}
\vskip -0.4in
\end{figure}

% \subsection{Zero-Shot Capabilities}
% \label{app:zero-shot}

% \begin{table*}[h]
% \caption{Accuracy Generalization to Unseen WorldsTODO}
% \label{table-zeroshot}
% \vskip 0.15in
% \begin{center}
% \begin{small}
% \begin{sc}
% \begin{tabular}{lcccccr}
% \toprule
%  Model & Angle  & Weight & Speed & Height/Width  \\
% \midrule
% Random Control & & &         \\
% \midrule
% GPT-3 davinci & & &         \\
% GPT-3 instruct davinci & & &         \\
% GPT-3 babbage & & &         \\
% GPT-3 curie & & &         \\
% GPT-2 & & &         \\
% \bottomrule
% \end{tabular}
% \end{sc}
% \end{small}
% \end{center}
% \vskip -0.1in
% \end{table*}

%%%%%%%%%%%%%%%%%%%%%%%%%%%%%%%%%%%%%%%%%%%%%%%%%%%%%%%%%%%%%%%%%%%%%%%%%%%%%%%
%%%%%%%%%%%%%%%%%%%%%%%%%%%%%%%%%%%%%%%%%%%%%%%%%%%%%%%%%%%%%%%%%%%%%%%%%%%%%%%

\end{document}